# VirtualTrialHarness: Coordinating Agentic Operationalization of Virtual Trial Protocols Across Complex Clinical Schemas

Pengze Li, Jingna Feng, Yue Yu, Yining Wu, Xinyue Hu, Jie Lin, Avanti Bhandarkar, Cui Tao

AI & Informatics, Mayo Clinic, Jacksonville, FL, USA

Correspondence: Cui Tao (Tao.Cui@mayo.edu)

## Abstract

Virtual trials use routinely collected clinical data to evaluate treatment strategies, but translating a study protocol into an executable analysis requires determining how clinical concepts and study requirements should be represented in the available data and applied consistently across the workflow. These decisions depend on local data representations and operational conventions and can involve source selection, patient and encounter identity, linkage, and temporal interpretation. Here we introduce VirtualTrialHarness (VTHarness), an agent framework that coordinates virtual-trial operationalization by selecting reusable procedural skills for each workflow stage and uses explicit contracts to maintain consistent study decisions across stages. We also introduce VTBenchmark for evaluating source selection and cohort construction. Across three complex synthetic environments, VTHarness successfully produced all required cohort outputs matching the references in 95.1% of workflows after requirement clarification, compared with 60.4% using retrieval from the same procedural corpus. Procedural skills improved source recovery across three language models and increased source overlap with analyst code in 13 Mayo Clinic cases. Site-specific adaptation enabled reference-matched cohort execution across eight MIMIC-IV tasks. Within the Mayo Clinic Virtual Trials Program, the framework supported investigator review of cohort construction and endpoint analysis. These findings support coordinating procedural guidance with workflow decisions to improve virtual-trial implementation across complex clinical schemas.

## 1 Introduction

Randomized controlled trials (RCTs) are the benchmark for evaluating clinical interventions, but recruiting and following participants can be costly or infeasible [1]. Clinical data recorded in hospital electronic health records (EHRs) can complement RCTs with evidence on effectiveness, safety, and treatment-effect heterogeneity [2]. We use *virtual trial* (VT) to denote a protocol-defined treatment comparison implemented using routinely collected EHR data. Target-trial emulation specifies the protocol of a hypothetical randomized trial—including eligibility criteria, treatment strategies, time zero, follow-up, outcomes and estimand—and implements it using observational records [3]. Transparent reporting also requires specifying how the target-trial protocol is mapped to the observational data used for its emulation [4]. Our focus is the operational translation of a supplied protocol into linked cohort and analysis workflows.

Implementing a VT protocol requires mapping clinical requirements to site-specific data structures, vocabularies and operational conventions [5,6]. Common data models such as OMOP standardize data structures and clinical concepts, substantially reducing heterogeneity [7,8]. However, standardization does not guarantee that every study-relevant source or clinical detail has been mapped, nor does it eliminate local differences in linkage, data coverage, and event interpretation [9,10]. Implementation may therefore require working across standardized, partially harmonized, and native hospital data. Even when relevant records conform to a common model, their completeness and suitability for the intended analysis still require assessment [11].

Operationalizing virtual trials in health-system data environments can be substantially more complex than working with curated research datasets. Research datasets such as MIMIC-IV undergo preparation that consolidates and reorganizes source tables to facilitate retrospective analysis [12]. In institutional environments, by contrast, clinically related information may be distributed across operational, terminology, and specialty data sources, with multiple representations of patients, encounters, clinical events, and event times. In our work with physician-submitted VT requests at Mayo Clinic, for example, the institutional data dictionary covered more than 150 tables and 2,000 documented fields in its base EHR component alone, with additional specialty datasets (Supplementary Section S1). Although an individual study may use only a small subset of this catalog, identifying that subset and determining how its records should be linked and interpreted requires knowledge of local data practices. Enterprise text-to-SQL benchmarks such as Spider 2.0 similarly emphasize metadata search and multi-step execution across complex database environments [13].

These challenges make VT operationalization more than a criteria-to-SQL task. Electronic phenotyping and criterion-to-query systems address important parts of the problem, including concept extraction, temporal interpretation, and executable cohort definition [14-16]. Formal representations such as Temporal Ensemble Logic additionally capture temporal dependencies across trial components [17]. Yet identifying candidate fields is different from determining how they should be used at a particular hospital. Schema metadata describes available fields and relationships, but does not by itself establish which source, linkage path or event clock satisfies a particular protocol. For example, implementing a treatment-initiation rule may require distinguishing prescription orders from administrations, following a locally validated patient-linkage path, and selecting an event time that consistently anchors eligibility and follow-up. A query can execute successfully while using the wrong treatment source or temporal anchor. Resolving these choices often depends on knowledge accumulated through prior analyst implementations and local operational documentation [10]. VT operationalization therefore requires local procedural knowledge to guide individual queries and remain consistent across the workflow.

Agents built on large language models (LLMs) offer a way to combine protocol interpretation, access to local resources, and iterative execution [18-20]. Recent systems support study-to-code translation, multi-stage EHR analysis, and reuse of prior analytic experience [21-23]. Clinical retrieval-augmented generation studies also show that corpus organization and retrieval configuration influence performance on medical knowledge tasks [24,25]. However, access to procedural documentation alone does not determine which procedure applies, when it is needed or how an accepted decision should constrain later stages. These responsibilities fall to the *agent harness*, the software layer that manages an LLM's context, tools, workflow state and checks [26-29]. A central question is whether coordinating the selection and exposure of local procedures with the current workflow stage improves protocol execution beyond query-driven retrieval of the same procedural corpus.

We introduce VirtualTrialHarness (VTHarness), an agent harness with a role-based multi-agent architecture for VT operationalization. Specialized roles cover protocol interpretation, source grounding, cohort-base construction, eligibility assessment and endpoint analysis. In the evaluated implementation, a single language-model agent enacted these roles sequentially. Reusable *skills* express recurring source, identity and event-time rules separately from study-specific clinical definitions. Progressive selection and exposure use the current requirement, workflow stage and acquired schema information to choose applicable skills and supply their operation-specific guidance. Explicit contracts specify how each role should use accepted study decisions and which outputs and checks it must provide. Schema tools and execution feedback support implementation, while human review allows study outputs to be inspected and revised. Within the Mayo Clinic Virtual Trials Program, physicians and analysts reviewed how VTHarness interpreted and implemented study requirements.

Existing benchmarks assess clinical text-to-SQL question answering, as in EHRSQL, and interactive patient-level EHR tasks, as in MedAgentBench [30,31]. Evaluating VT operationalization additionally requires assessing whether agents select appropriate data sources and preserve protocol requirements across interdependent executable outputs. We therefore also introduce VTBenchmark, a dedicated benchmark for VT operationalization. Its execution environments abstract challenges encountered in institutional VT implementation: selecting among clinically related sources, resolving patient and encounter identities, interpreting event times and navigating a broad catalog. Its complex-schema component comprises 24 base protocol tasks, each represented in four synthetic database environments that capture these challenges. Its source-grounding component comprises 90 study tasks derived from target-trial emulation publications and Mayo Clinic intake materials and assesses recovery of the data sources needed to implement study requirements. Together, these components assess source grounding and executable workflow delivery.

Using VTBenchmark, we test whether the way procedural knowledge is selected and supplied changes implementation outcomes. In the three complex-schema environments, progressive skill selection and exposure achieved 95.1% complete delivery of the five required cohort products after clarification of a study requirement, compared with 60.4% for query-driven retrieval of the same procedural corpus within the same harness. Additional evaluations examined skill use across model backbones and Mayo Clinic cases, and site adaptation in MIMIC-IV. Together, the framework and benchmark assess how procedural guidance supports correct cohort construction and consistent summaries.

## 2 Results

### 2.1 Contract-coordinated roles supported cohort and analysis workflows

We applied VTHarness within the Mayo Clinic Virtual Trials Program to support VT intake requests submitted by clinical investigators. VTHarness combined five specialized agent roles with a shared harness for procedural guidance, study state and execution checks (Fig. **1**). The roles covered protocol interpretation, source grounding, base-cohort construction with treatment assignment and time zero, eligibility and attrition, and endpoint analysis. In the evaluated implementation, one LLM agent enacted these roles sequentially. Contracts specified how each role should use previously accepted decisions and outputs. The shared harness selected relevant local skills and supplied operation-specific guidance alongside schema information. This structure linked protocol requirements and selected data sources to cohort records, eligibility flags, attrition summaries and analysis specifications.

Within this institutional workflow, VTHarness supported cohort construction, propensity-score and balance diagnostics [32,33], and time-to-event analysis, including Kaplan–Meier curves [34], Cox regression [35] and competing-risk analysis [36]. Reviewed cohorts and analysis specification provided the basis for these analyses, with study decisions and results documented together.

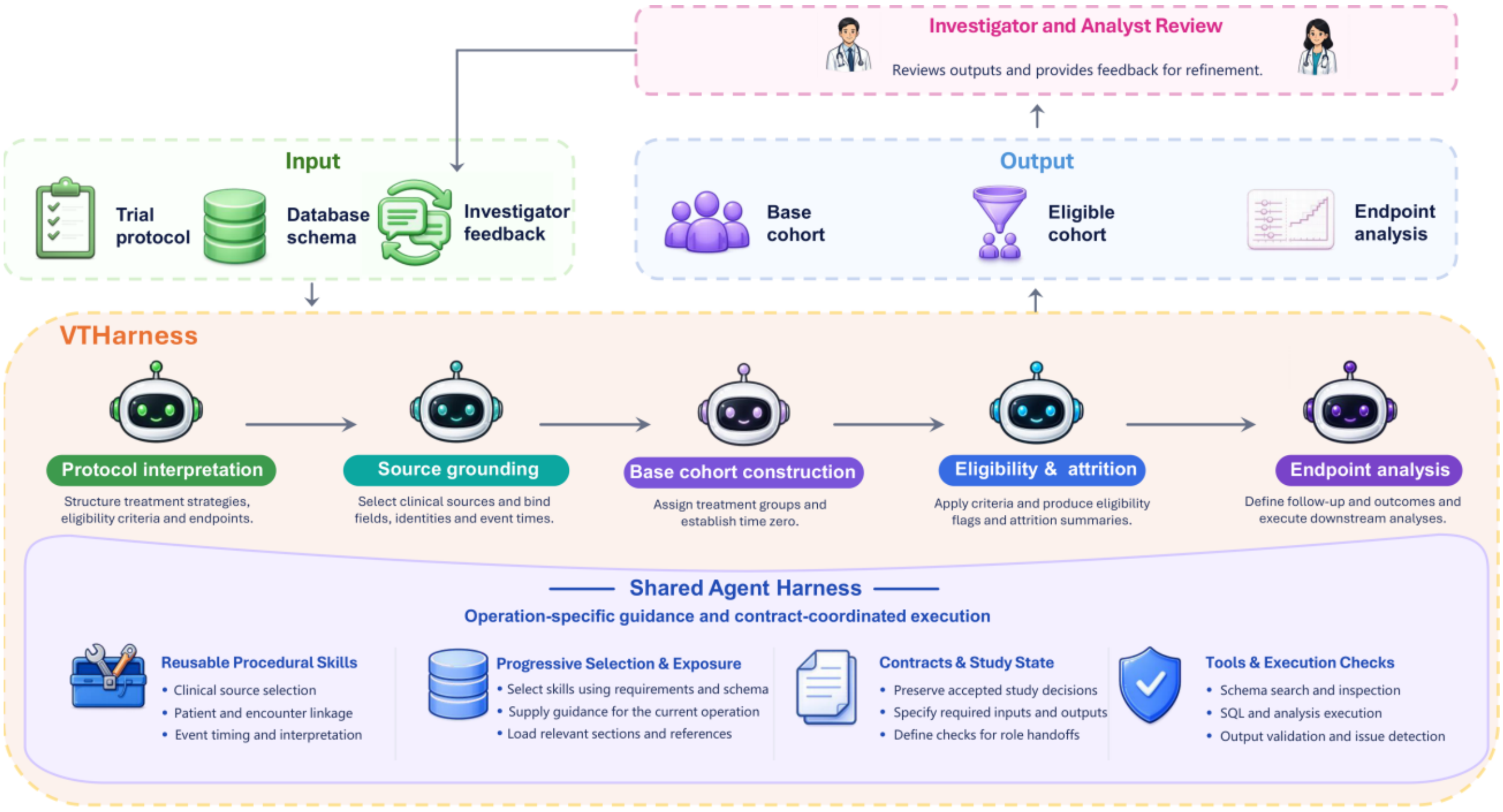

**Figure 1. VTHarness architecture and contract-coordinated workflow.** VTHarness comprises the five specialized agent roles and the shared agent harness enclosed by the orange boundary. Trial protocols, database schema and investigator feedback guide cohort construction and endpoint analysis. The shared harness selects reusable procedural skills and supplies operation-specific guidance using study requirements, schema information and workflow state. Contracts preserve accepted decisions and define required inputs, outputs and handoff checks. Tools support schema inspection, code execution and output validation. Investigators and analysts review outputs and provide feedback for revision. In the evaluated implementation, a single language-model agent enacted the roles sequentially using a shared backbone.

The interface allowed investigators to inspect cohort results and analysis specifications before cohort confirmation and during endpoint analysis (Fig. **2**). The illustrated interface uses synthetic demonstration data.

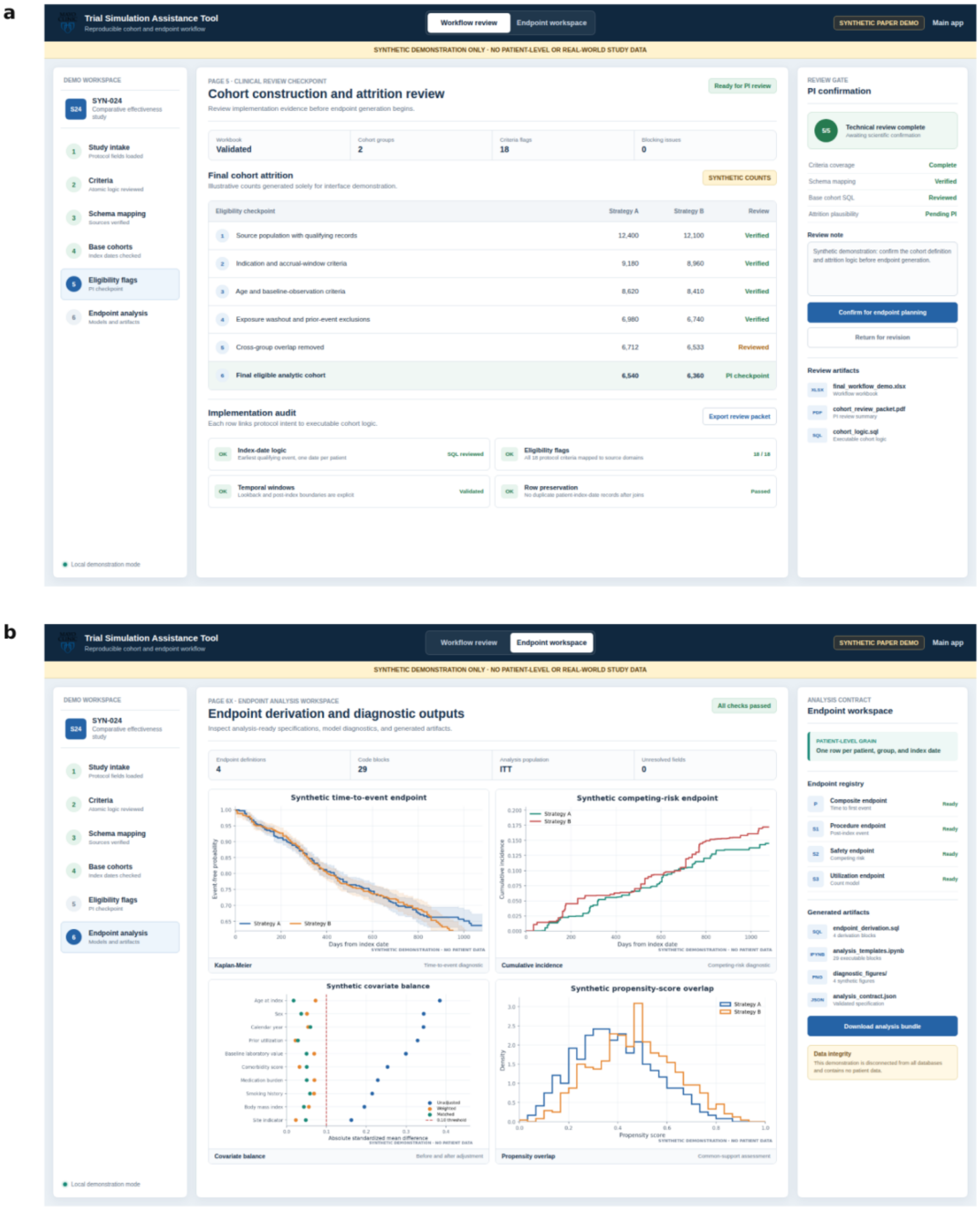

**Figure 2. Cohort review and transition to endpoint analysis. a,** The cohort-review panel shows workflow progress, arm-specific attrition, consistency checks and cohort results for investigator review and revision. **b,** After cohort confirmation, the analysis workspace shows the analysis unit, endpoint definitions, generated outputs and four types of diagnostic display. Both panels use synthetic demonstration data. Displayed counts and curves are illustrative.

## 2.2 VTBenchmark assessed source grounding and cohort execution

We constructed VTBenchmark to assess two linked aspects of VT operationalization: selecting appropriate data sources and delivering executable cohort products. Its execution component represented challenges encountered in institutional workflows, including alternative clinical evidence sources, identity linkage, temporal interpretation and catalog navigation. The complex-schema component comprised 24 base protocol tasks (protocol roots), each represented across four synthetic database environments, yielding 96 task–environment instances. The 24 roots covered all combinations of four clinical contexts and six operational challenges (Methods, Section 4.3.1). The environments were Simple reference (clear sources and short linkage paths), Source ambiguity (competing clinical sources), Temporal and representation ambiguity (event-time and data-representation choices), and Large-catalog navigation (a 1,024-table catalog with multi-hop joins). Each root retained its clinical semantics across these environments. Each task required five linked products: patient–group–index base records, condition-level flags, final eligibility, single-condition counts and cumulative attrition. Complete delivery required successful execution and agreement with the references for all five outputs under the evaluation procedure described in Methods (Section 4.6.1).

The source-grounding component comprised 90 constructed study tasks based on 55 target-trial emulation publications and 35 institutional virtual-trial intake requests at Mayo Clinic, represented in a 1,024-table catalog. It assessed recovery of expected source tables for eligibility criteria and treatment arms. Together, the two components assessed source grounding and complete cohort delivery. Additional evaluations outside VTBenchmark examined13 historical VT cases with corresponding analyst notebooks, eight MIMIC-IV adaptation tasks and an institutional case study.

## 2.3 Progressive skill selection and exposure improved cohort delivery

We compared four experimental configurations of supplementary knowledge support within the same harness, keeping the workflow, language model, tools, requirement clarification and execution budgets constant. Schema-only support provided the task, schema tools and generic tool instructions without supplementary knowledge-library resources. Raw-evidence retrieval-augmented generation (RAG)[37] retrieved descriptive resources, including data-dictionary entries, field descriptions, relationship metadata, value domains and source documents, but excluded procedural modules and worked examples. Shared procedural-corpus retrieval used query-driven BM25 [38] retrieval over all available knowledge resources, including procedural skills. Full VTHarness used the same resources, selecting and presenting relevant procedures according to the current workflow stage and operation. For each task, all configurations received the same predefined clarification of one upstream requirement, such as the treatment source or time anchor, before revising the initial implementation. Post-feedback refers to the resulting revised phase.

Across the three complex environments (Source ambiguity, Temporal and representation ambiguity, and Large-catalog navigation), progressive skill selection and exposure achieved complete post-feedback delivery in 137/144 workflows (95.1%), compared with 87/144 (60.4%) for query-driven retrieval of the same admissible procedural corpus (Table **1**; Fig. **3**). The paired difference was 34.7 percentage points (Bonferroni-adjusted 97.5% root-bootstrap confidence interval, 25.0 to 43.8). Against schema-only support, the difference was 95.1 percentage points (Bonferroni-adjusted 97.5% confidence interval, 91.7 to 98.6). Schema-only support and raw-evidence RAG completed no revised workflows in these environments. The shared-corpus contrast demonstrated a delivery advantage for operation-specific selection and exposure over query-driven retrieval.

Here, progressive skill selection and exposure refer to selecting applicable skills from task interpretation and acquired schema information, then loading the operation-specific sections and referenced procedures needed for the current action. Cohort-base construction and eligibility assessment used the corresponding procedural guidance. Clarified study requirements were incorporated before the implementation was revised. The two procedural configurations therefore shared the available corpus, while the content selected and presented to the model could differ (Methods, Section 4.2.2).

Each of 24 protocol roots was represented in four synthetic schema environments: Simple reference with 32 relations, Source ambiguity with 128 relations, Temporal and representation ambiguity with 128 relations, and Large-catalog navigation with 1,024 relations and multi-hop joins. The environments combined variations in source choice, linkage structure, event-time interpretation and catalog scale. For each root, environment and knowledge configuration, two technical repeats (separate end-to-end workflow executions of the same constructed task) yielded 192 workflows per configuration and 768 in total. Each workflow included initial implementation and revision after clarification of a study requirement. Programs were evaluated using the complete-delivery criteria described in Methods (Section 4.6.1).

**Table 1. Complete cohort delivery in the four-arm knowledge experiment.**

| Knowledge configuration | Initial | Post-feedback | Complex post-feedback |
|---|---|---|---|
| Schema-only | 32/192 | 33/192 | 0/144 |
| Raw-evidence RAG | 23/192 | 13/192 | 0/144 |
| Shared procedural-corpus retrieval | 140/192 | 132/192 | 87/144 |
| Progressive skill selection and exposure (VTHarness) | 173/192 | 181/192 | 137/144 |

All four configurations used the same VTHarness workflow. The final row denotes its full selection-and-exposure policy. Initial and post-feedback columns include all four environments (192 assigned workflows per configuration); complex post-feedback includes the three complex environments (144 workflows per configuration). The two procedural configurations had access to the same corpus. For the pooled complex-environment contrasts, technical repetitions and environments were averaged within each of 24 root cases before root-level bootstrap resampling. All assigned workflows contribute to the reported proportions.

In the Simple reference environment, procedural retrieval and progressive skill selection and exposure performed similarly, completing 45/48 and 44/48 revised workflows, respectively. Across all four environments, revision changed 19 VTHarness workflows from incomplete or incorrect to complete correct delivery, while 11 changed in the opposite direction, yielding a net increase from 173/192 to 181/192 complete deliveries. Systems could regenerate all affected cohort outputs after feedback. Environment-specific counts and transitions are reported in Supplementary Tables S2 and S3.

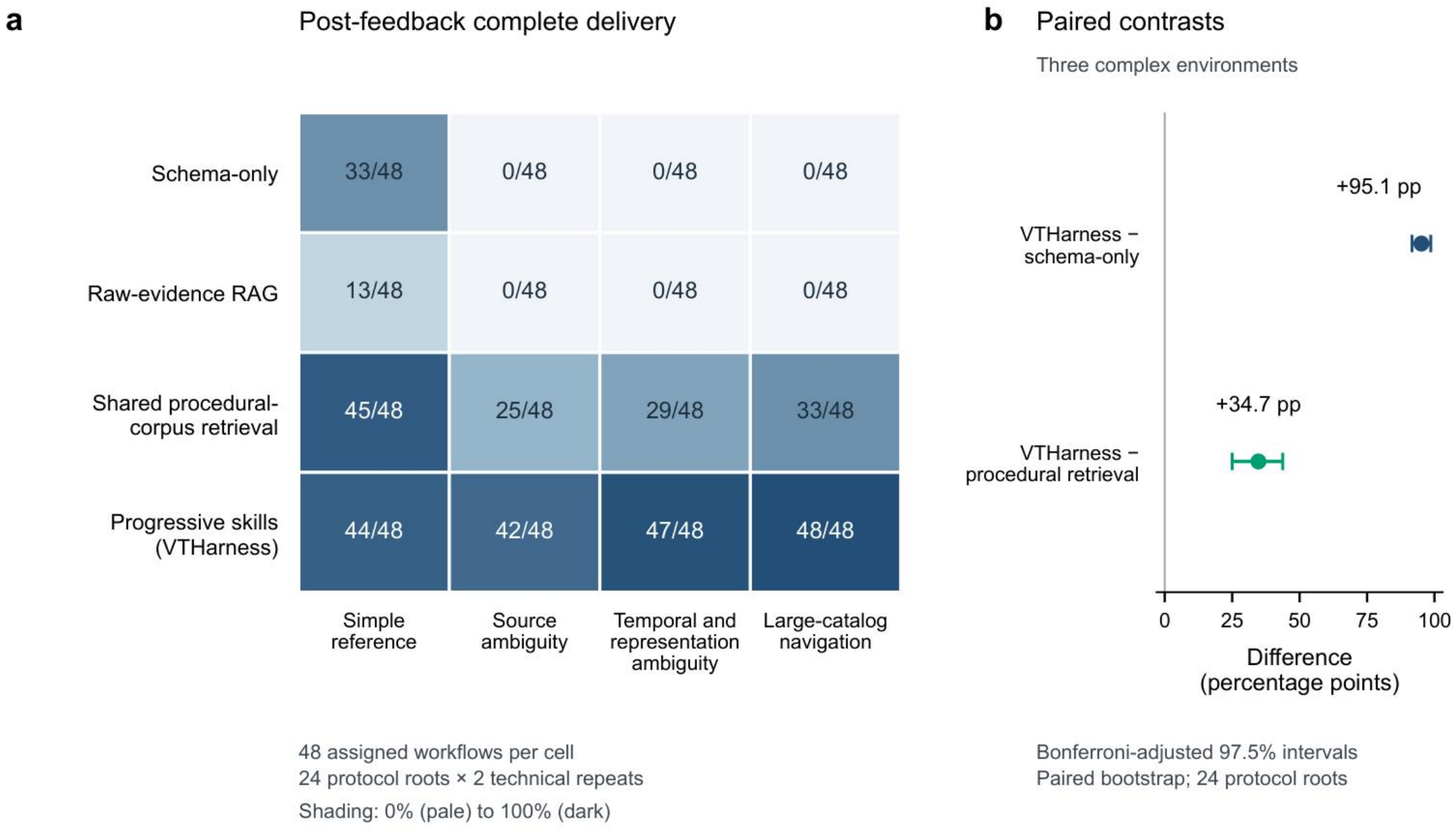

**Figure 3. Knowledge-use policies and complete cohort delivery. a,** Observed post-feedback complete deliveries in each configuration and environment; each cell contains 48 workflows (24 roots and two technical repeats). Shading encodes the completed proportion, with exact counts printed in each cell. The heat-map columns appear from left to right as Simple reference, Source ambiguity, Temporal and representation ambiguity, and Large-catalog navigation. Multiple structural and semantic properties varied together across environments. **b,** Prespecified paired differences pooled across the three complex environments with Bonferroni-adjusted 97.5% root-bootstrap confidence intervals. Repeats and environments were averaged within each of the 24 resampled roots. The VTHarness label denotes progressive skill selection and exposure; all four configurations used the same harness workflow. VTHarness and procedural retrieval had access to the same admissible corpus. Their contrast assesses the combined selection-and-exposure policy, including content selection, presentation order and context volume.

## 2.4 Aligned program execution with correct cohort delivery

We also compared VTHarness with adapted versions of EmulatRx and HealthFlow [22,23] in the Large-catalog navigation environment. Each system was assigned the same 24 protocol roots with two technical repetitions, yielding 48 workflows per system. All systems had access to the same procedural corpus while retaining their own knowledge-use and execution mechanisms. We assessed three indicators: successful program execution, assessability of all five cohort outputs and complete correct delivery (Methods, Section 4.6.1). Assessability required all five outputs to be readable and mappable to predefined cohort structures. For the adapted systems, empty base cohorts were classified as unassessable. Complete correct delivery additionally required successful execution and agreement with the references for all five outputs.

After feedback, VTHarness achieved successful execution in 46/48 workflows, five-product assessability in 46/48, and complete correct delivery in 44/48 (Fig. **4**). The corresponding outcomes were 41/48, 18/48 and 1/48 for adapted HealthFlow, and 23/48, 3/48 and 0/48 for adapted EmulatRx. For the evaluated adaptations, successful program execution was more frequent than delivery of a complete, reference-correct cohort package. These measures distinguished whether programs ran successfully, whether all required outputs could be evaluated and whether those outputs matched the references.

Output analysis showed that successful execution could coexist with errors in implementing cohort requirements. Among workflows that executed successfully but did not yield five assessable products, all 23 HealthFlow workflows and 20 EmulatRx workflows returned empty base cohorts despite nonempty reference cohorts. Examination of individual cases traced incorrect cohort results to errors in patient identity linkage and failure to apply a revised source requirement.

Initial and revised per-product results are reported in Supplementary Tables S4 and S5. Results for the four-environment comparison, which used different knowledge inventories, are provided in Supplementary Table S6.

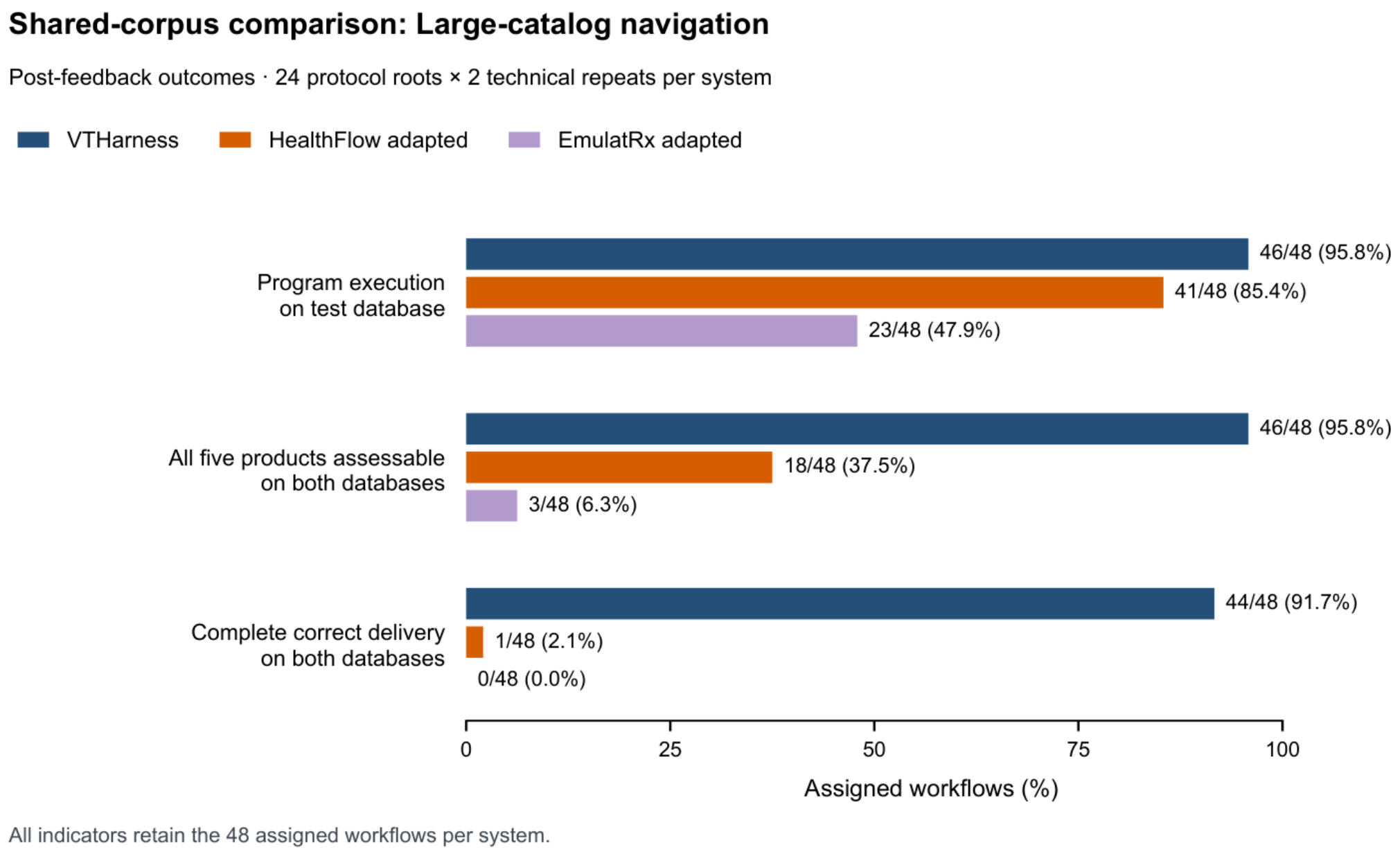

**Figure 4. Program execution, product assessability and complete correct delivery in the Large-catalog navigation environment.** Each indicator uses all 48 assigned workflows per system (24 protocol roots with two technical repetitions); labels report counts and percentages. Program execution denotes successful execution of the submitted program on the test database. Five-product assessability requires all five explicit outputs to be readable and mappable to predefined cohort structures on both databases, without requiring reference agreement. For the adapted systems, empty base cohorts were classified as unassessable. Complete correct delivery additionally requires reference agreement for all five products after normalization and successful execution. All systems had access to the same procedural corpus. Comparisons are descriptive, with technical repetitions nested within protocol roots. Source Data report initial, revised and per-product outcomes.

## 2.5 Skills improved source recovery across language models

To assess procedural grounding across language-model backbones, we evaluated source selection in VTBenchmark's 90-study component using a 1,024-table catalog. The constructed study tasks were based on 55 target-trial emulation publications and 35 institutional virtual-trial intake requests at Mayo Clinic. Each backbone was evaluated in 540 assigned episodes, comprising 90 studies, two configurations of the same harness with and without skills, and three technical repetitions. Expected-table recall measured recovery of reference source tables for eligibility criteria and treatment arms, with missing or invalid outputs receiving zero source-recovery credit. Scores combined criterion- and arm-level requirements, with repetitions averaged within each study and configuration before paired study-level bootstrap resampling.

Skills improved unconditional expected-table recall for all three backbones (Fig. **5**). GPT-5.1 [39] increased from 0.489 to 0.636, a paired difference of 0.146 (95% paired study-bootstrap confidence interval, 0.118 to 0.175); Qwen3-30B-A3B [40] increased from 0.113 to 0.444, a difference of 0.331 (0.297 to 0.365); and MedGo [41] increased from 0.049 to 0.180, a difference of 0.131 (0.096 to 0.167). F1, F2 and complete-route outcomes are reported in Supplementary Table S7.

Skills also increased the proportion of source-selection outputs that met the requirements for scoring. With versus without skills, delivery was 99.3% versus 97.0% for GPT-5.1, 95.9% versus 35.9% for Qwen3-30B-A3B, and 51.1% versus 22.2% for MedGo. The primary recall estimates retained all assigned episodes, so they reflected both output delivery and source recovery. Supplementary Fig. S1 reports delivery rates and descriptive source recall among delivered outputs. For MedGo, improved source recovery was accompanied by more recorded model and tool calls (Supplementary Table S8).

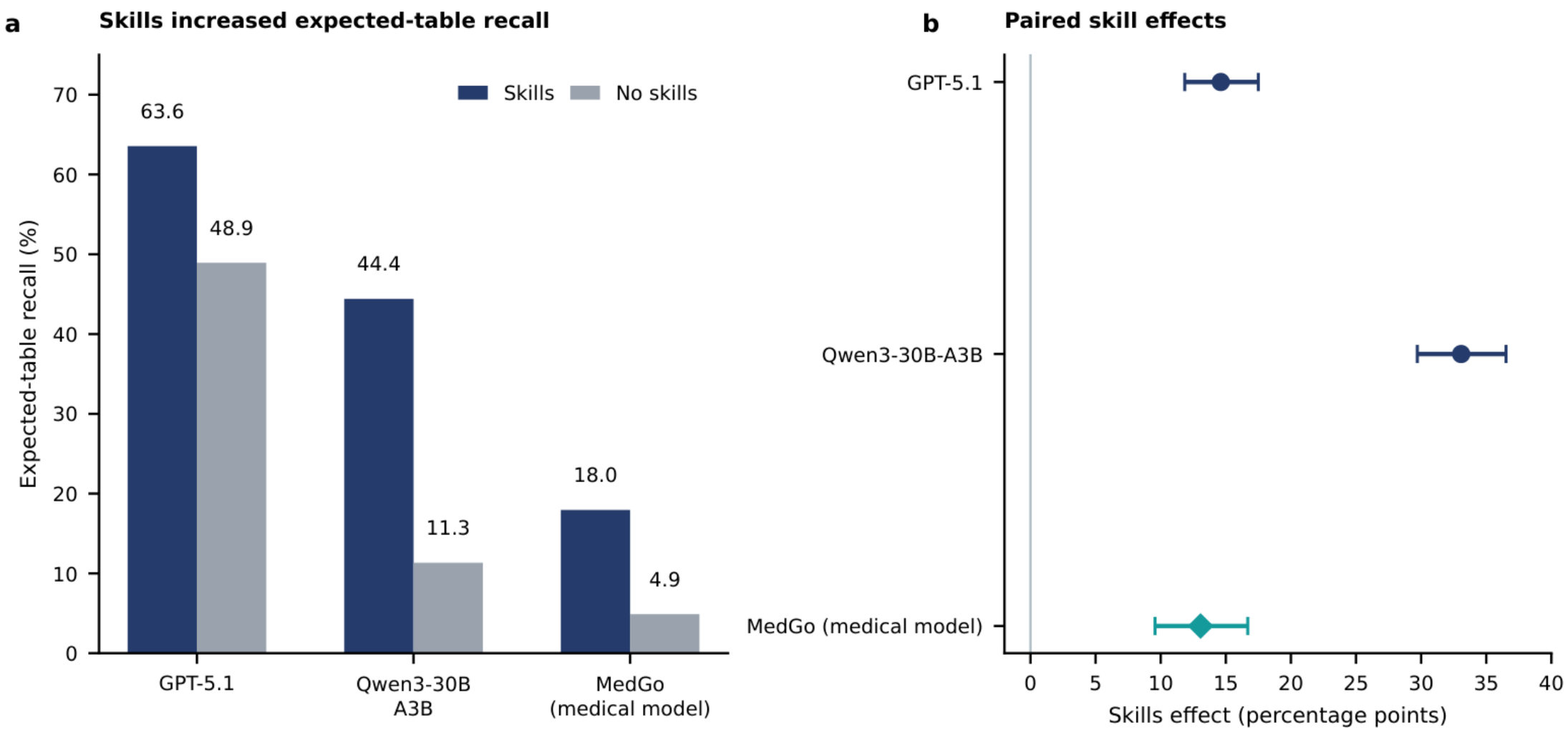

**Figure 5. Procedural skills improved source recovery across three language-model backbones. a,** Unconditional expected-table recall for the same staged harness with and without skills. Missing or invalid outputs received zero source-recovery credit. Each backbone was evaluated in 270 episodes per configuration (90 studies and three technical repetitions), yielding 540 episodes in total. Scores combined criterion- and treatment-arm-level requirements; repetitions were averaged within each study and configuration, and studies received equal weight. **b,** Paired recall differences between the skill and no-skill configurations, with 95% paired study-bootstrap confidence intervals. Differences are with skills minus without skills, and studies were resampled separately for each backbone. MedGo is marked as the medical-model backbone. These outcomes measure recovery of the expected source tables for eligibility criteria and treatment arms.

## 2.6 Skills increased source overlap with analyst implementations

We next examined whether procedural skills improved source selection for 13 Mayo Clinic VT requests. We compared sources selected by VTHarness with those identified in the corresponding analyst code.The notebooks were withheld from VTHarness, which was evaluated with and without skills using three repetitions per case. Reference tables were identified in cohort code and comments, and reference fields were extracted from SQL queries.

With skills, recall of tables identified in analyst code was 60.7%, compared with 26.6% without skills. The paired difference was 34.2 percentage points (95% paired study-bootstrap confidence interval, 20.9 to 48.4; Fig. 6). Recall of table-qualified fields was 52.5% versus 21.8%, a difference of 30.7 percentage points (18.8 to 44.4). Repetitions were averaged within each case and configuration, and missing outputs received zero source-recovery credit.

The notebooks reflected an iterative clinical workflow in which physicians refined requests and analysts revised code. These results therefore indicate closer correspondence with existing implementations. The separate, author-adjudicated measure of requirement representation showed no consistent improvement (Supplementary Table S9).

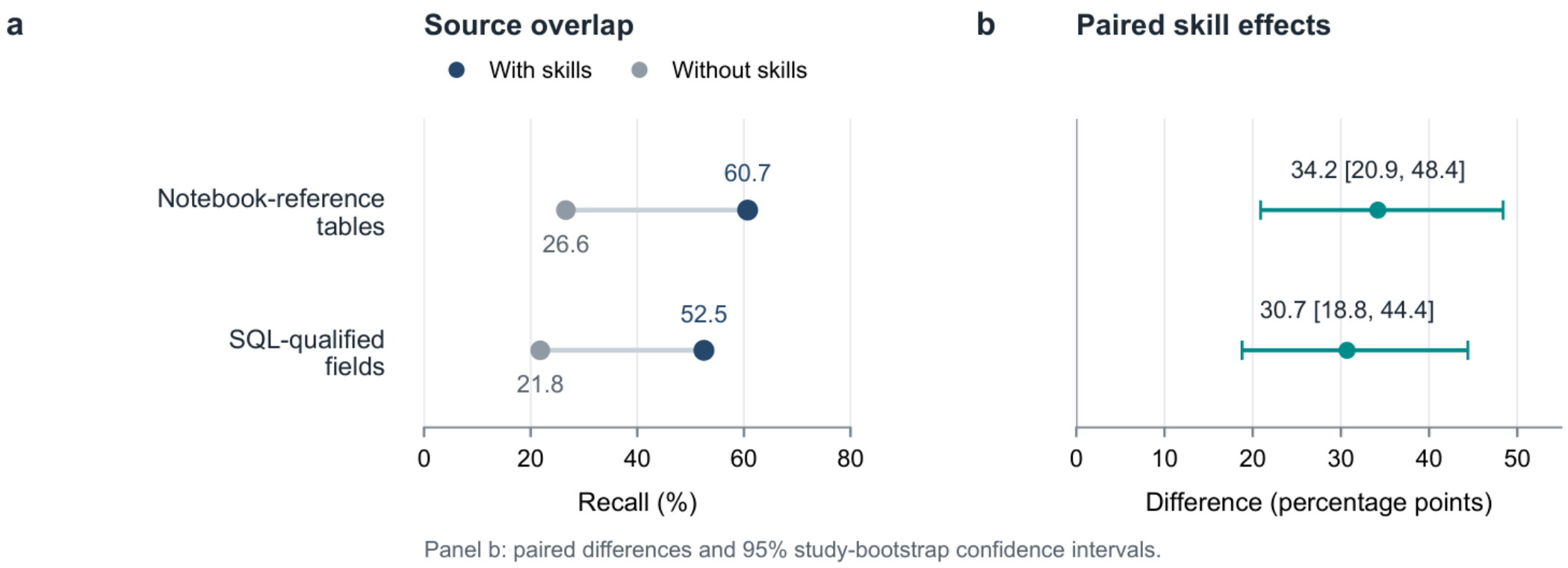

**Figure 6. Source overlap with analyst implementations in 13 Mayo Clinic VT cases.** VTHarness received the study requests, while analyst notebooks provided the source references. Reference tables were identified by matching data-dictionary names in selected cohort code and comments. Reference fields were extracted from parseable SQL and validated against the schema, retaining their table assignments. Evaluated sources comprised the tables and fields explicitly selected by VTHarness. **a,** Recall with and without skills, after averaging three repetitions within each case and configuration. **b,** Paired recall differences in percentage points, with 95% paired study-bootstrap confidence intervals. Differences were computed before rounding. Missing final source outputs received zero credit; empty reference field sets remained unscored. The auxiliary author-adjudicated requirement-representation outcome was assessed separately.

## 2.7 Site-specific adaptation enabled MIMIC-IV cohort execution

To assess adaptation across databases and task sets, we applied VTHarness to MIMIC-IV 3.1 [12,42]. The evaluation used eight cohort-construction tasks over nine tables from a deterministically selected subset of 1,000 patients. Tasks covered demographic eligibility, laboratory measurements, intensive care unit (ICU) duration and observations, discharge status and medication administration. We encoded MIMIC-IV conventions for patient, hospitalization and ICU identities, admission-based age, event timing, measurement units and medication administration in five reusable skills. We also adapted the execution procedures to the database. The harness selected the relevant skill content according to the data sources and current workflow stage. For example, medication administration evidence was linked to the selected hospitalization and time window, whereas ICU observations were linked to the selected stay. Site procedures were reused where applicable, with thresholds and time windows supplied by individual task specifications.

After developing MIMIC-IV-specific skills and execution procedures, we ran the resulting configuration once on each of the eight development tasks. All eight runs produced the five required cohort outputs in exact agreement with the references. These included full index timestamps, condition flags for every base patient and cumulative attrition (Supplementary Tables S10 and S11).

### 2.8 Analysts valued reviewability and estimated time savings

To characterize user experience and perceived workload, we collected structured questionnaire responses from five analysts who had used VTHarness in the institutional workflow. Respondents estimated the active work required with and without the system for a common VT scenario, assuming immediate physician responses to clarification questions. All five estimated less active work with VTHarness, although the magnitude of the estimated reduction varied across respondents (Fig. **7**a). Median estimated active work was 40 hours without the system and 28 hours with it. Calculating savings within each respondent before summarizing across respondents yielded a median reduction of 25 hours (95% respondent-bootstrap confidence interval, 6 to 60). The median within-respondent percentage reduction was 37.5%.

Ratings emphasized reviewability and support for working with local data (Fig. **7**b). On the five-point scales, median ratings were 5 for review appropriateness, auditability, local grounding, visibility of unresolved data-source choices, future reuse and review efficiency. Confidence and skill value had median ratings of 4, while unnecessary complexity had a median rating of 2. All five respondents selected protocol decomposition, schema mapping and staged criteria application among the most useful workflow components (Fig. **7**c). These responses identified the organization and inspection of intermediate study decisions as valued aspects of the workflow.

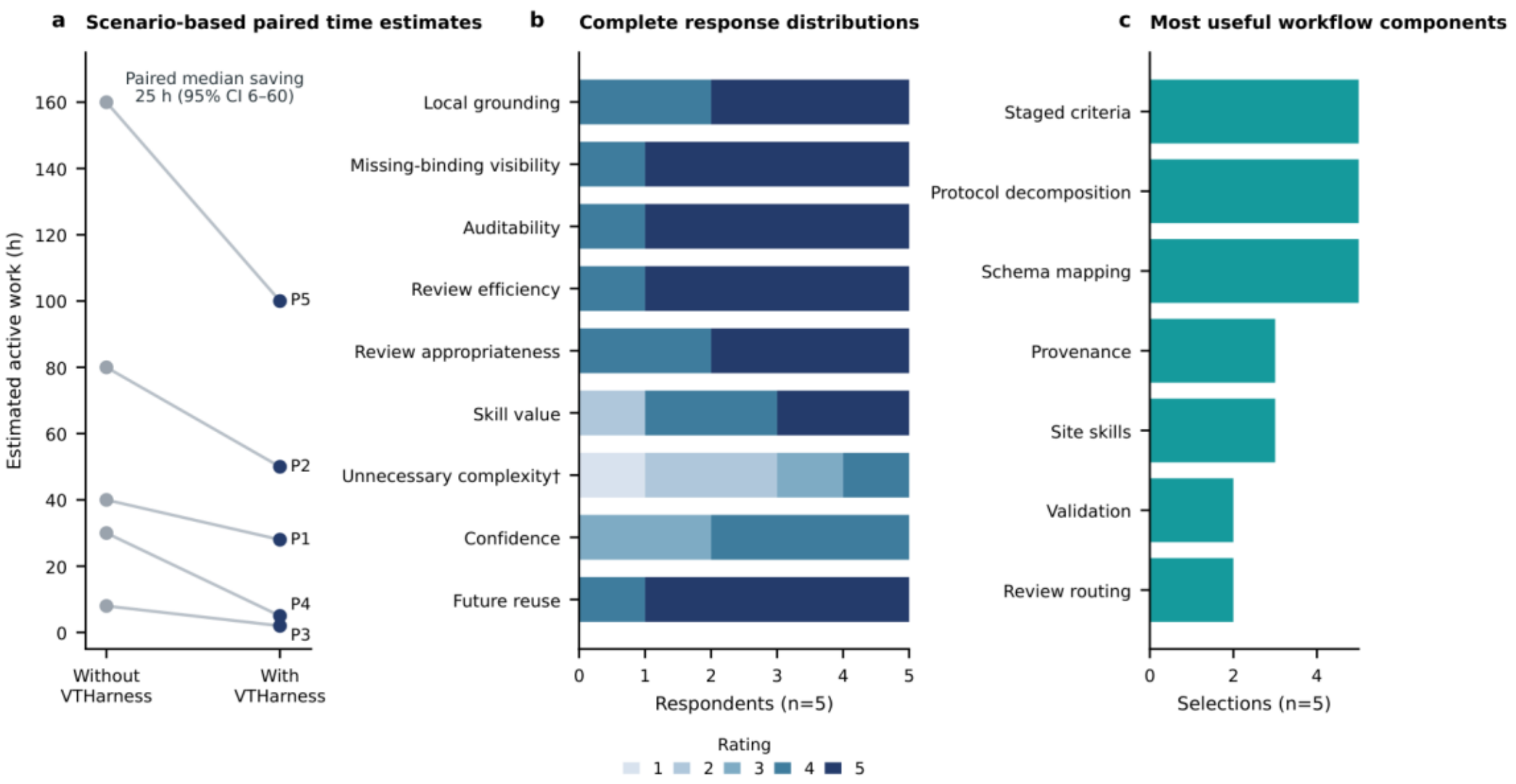

**Figure 7. Analyst feedback on workload and workflow usefulness.** Five analysts with experience using the institutional workflow completed the questionnaire. **a,** Paired estimates of active work with and without VTHarness for a common VT scenario, assuming immediate physician clarification. Each line connects estimates from the same respondent. The annotation reports the median within-respondent saving and its 95% respondent-bootstrap confidence interval. Times are scenario-based estimates. **b,** Complete response distributions for the nine five-point questionnaire items. A lower rating indicates less unnecessary complexity for the reverse-direction item marked by a dagger. **c,** Numbers of respondents selecting each workflow component among the most useful; respondents could select multiple components.

## 3 Discussion

Coordinating procedural knowledge with workflow decisions improved the implementation of virtual-trial requirements in complex synthetic schemas. The controlled within-harness experiment provides the primary evidence for the benefit of progressive skill selection and exposure. This policy increased complete cohort delivery over query-driven retrieval of the same procedural corpus. This finding supports selecting guidance for the operation being performed as a consequential part of agentic workflow design. VTHarness expresses recurring local data operations as skills and uses workflow state to select the guidance supplied at each decision. Contracts preserve agreed choices of data sources, patient and encounter identities, and time anchors across successive stages. VTBenchmark evaluates whether the resulting source selections and cohort outputs satisfy study requirements.

The clinical importance of this connection follows from the dependency structure of target-trial emulation. Treatment assignment and time zero must align with eligibility, baseline assessment, follow-up and censoring [3,43]. The RCT-DUPLICATE study illustrates how reproducing trial design and measurement in observational data affects agreement with randomized evidence [44]. Local implementation choices therefore help determine the study actually conducted. Treating a prescription as medication administration, or linking an observation to the wrong hospitalization, can change treatment classification or cohort membership while still producing executable SQL. Coordination across roles therefore involves preserving the clinical meaning of accepted decisions as each stage transforms them into code and study products. Explicit decisions and their downstream outputs allow investigators to inspect how the intended protocol becomes an analysis. This emphasis on inspectable implementation is aligned with RECORD recommendations to report population-selection algorithms, data linkage and access to analytic code [45].

Schema complexity creates competing interpretations as well as a larger search space. Several sources or linkage paths may be structurally valid while representing different clinical events or units of observation. Procedural guidance can help resolve these alternatives when it is supplied at the decision where they matter. Prior studies show that language-model performance can depend on where information appears within a long context [46]. Reusable skill libraries have also supported the composition of previously acquired behaviors in embodied agents such as Voyager [47]. One plausible explanation for the observed benefit is that stage-specific exposure reduces the need to reconstruct local conventions from retrieved fragments and reconcile instructions for different operations. Similar performance between progressive exposure and procedural retrieval in the Simple reference environment is consistent with greater benefit when plausible implementation routes compete. This approach complements action verification in Mid-Harness and evidence-based revision in VeriHarness [28,29], linking procedural guidance to checks on its consequences. OntoCodex and AgentDS-BUS offer related examples of specialized task decomposition [48,49]; VT workflows additionally require shared source, identity and timing decisions to remain consistent across cohort products.

The cross-model results suggest that procedural guidance can support both task completion and source selection. GPT-5.1 usually produced scoreable outputs with or without skills, whereas Qwen3 and MedGo showed larger changes in output delivery. Overall source recovery therefore reflects both completion of the required steps and recovery of expected tables. Examining these components separately may help identify where a model needs support, whether in completing the workflow or in choosing among plausible sources. Their relative contributions may also depend on model and deployment characteristics.

VTBenchmark makes the distinction between executable code and coherent study products measurable. Source-grounding assessment tests data-source selection, while cohort-execution assessment checks both patient-level decisions and aggregate summaries against explicit references. Assessing patient-level products alongside aggregate summaries exposes errors that cohort counts alone could conceal: identical cohort sizes can arise from different patient memberships or index dates. This complements the clinical workflows developed in EmulatRx, HealthFlow, MedCP and Biomni-based trial emulation [22,23,50,51]. The shared-corpus comparison evaluated the adapted systems under VTBenchmark's common five-product delivery requirement. Its separate execution, assessability and correctness measures showed how the ability to run code differs from the ability to deliver all required study products correctly. Successful execution could still yield empty cohorts despite nonempty references, showing why evaluation must examine cohort content as well as whether code runs.

Source-grounding across language-model backbones and source overlap in Mayo Clinic cases addressed different questions. In the 13 Mayo Clinic VT cases, notebooks reflected physicians' evolving requests and subsequent analyst code revisions. Increased source overlap therefore indicates closer correspondence with existing implementations. The requirement-representation outcome showed no consistent improvement. These results separate two review questions: whether the clinical request has been represented appropriately and whether the selected data sources can implement it. Organizing review around these questions may focus attention on decisions that affect the intended study.

The MIMIC-IV exercise examined site adaptation through local procedural skills and changes to execution procedures. Identity, timing, unit and administration rules were reused across demographic, laboratory, encounter and medication tasks. Reference agreement across the eight development tasks confirmed execution after adaptation.Separating reusable data-operation rules from task-specific thresholds and time windows makes local conventions available across study requirements. Database adaptation then involves establishing the destination's conventions, supplying them at the relevant stage and checking the resulting products. Reviewed changes to those conventions can be maintained in the site skill library, consistent with prior work on computable phenotype representation and cross-institutional implementation [6,9].

Making local conventions explicit may also allow implementation knowledge to accumulate across studies. A rule established for patient linkage or event timing can be reviewed and reused while clinical thresholds and time windows remain study-specific. This could reduce repeated interpretation of the same local data practices. Reuse also creates a maintenance need: an incorrect or outdated rule could affect several studies. Tracking rule applicability and checking its validity after schema changes would help assess the durability of this approach.

Application within the Mayo Clinic Virtual Trials Program connects these capabilities to physician requests, analyst implementation and investigator review. Investigators could review selected sources, cohort definitions and attrition before endpoint analysis. The five experienced users valued this reviewability, auditability and local grounding. Some revisions introduced new errors, highlighting the need to check how changed study decisions affect cohort construction and subsequent analysis. Investigator review provides an opportunity to resolve ambiguity and assess these consequences. The revision results also suggest that reliability should be assessed as study requirements evolve. A change in treatment source or time zero can affect eligibility, baseline windows and subsequent analysis. Contracts make accepted decisions and their dependencies explicit, providing a basis for examining these consequences. Testing whether dependency-guided checks reduce errors after different types of clarification would connect workflow design to the iterative nature of clinical research.

The endpoint-analysis workflow carries reviewed cohort definitions, treatment assignments and time-zero anchors into follow-up and outcome analysis. Extending contracts to capture estimands, adjustment choices and diagnostic requirements could further connect study design with executable analysis.

Prospective workflow studies can quantify active work, physician clarification, rework and skill-maintenance effort during routine VT implementation. This emphasis on integration and human factors is consistent with DECIDE-AI [52]; the questionnaire identifies reviewability and perceived workload as useful priorities.

VTHarness connects reusable local procedures, workflow decisions and linked study products in a common framework for virtual-trial operationalization. The shared-corpus comparison supports selecting procedural knowledge for the current operation. The MIMIC-IV exercise illustrates how database-specific conventions can be expressed as reusable skills within the same system structure. VTBenchmark provides an accompanying evaluation of source grounding and coherent cohort delivery, while investigator review connects these products to endpoint-analysis workflows.

## 4 Methods

### 4.1 Study design and evaluation overview

We investigated whether reusable local procedures, selected for the current workflow stage and operation, improve the translation of virtual-trial requirements into coherent cohort products. The principal experiment varied the selection and presentation of knowledge within the same harness while keeping the language model, tools, revision allowance and required outputs constant. Its primary outcome was complete delivery after clarification of a study requirement, defined by successful execution and reference agreement for all five cohort outputs (Section 4.6.1). Protocol roots were the units of analysis, with database environments and technical repetitions nested within roots.

VTBenchmark comprised a complex-schema execution component and a source-grounding component. The execution component supported a four-configuration knowledge experiment and two separate comparisons with adapted systems. The source-grounding component compared the same harness with and without skills across three language-model backbones. Additional evaluations examined 13 historical Mayo Clinic VT cases, site adaptation in MIMIC-IV, an institutional workflow and feedback from five experienced analysts. These evaluations addressed source recovery, cohort execution, database adaptation and use in practice through their respective outcomes.

The VTHarness architecture, staged workflow, contracts, selection-and-exposure policy and routing logic were finalized before VTBenchmark was constructed and before experiment-specific skills were developed. VTBenchmark tasks, including the 24 protocol roots, schema challenges, counterexamples and observed failures, were not used to modify the system or its skills. Each experiment used this unchanged system with data-environment-specific skills. MIMIC-IV followed the same separation: the system remained unchanged, whereas MIMIC-IV-specific skills and runtime interfaces were developed using the eight reported tasks. That experiment therefore assesses execution after site adaptation rather than held-out transfer.

## 4.2 Procedural skills and contract-coordinated agent roles

### 4.2.1 Study representation and agent roles

VTHarness receives a protocol or structured intake, an authorized database catalog and reusable local procedures. Five stage-specific agent roles coordinate the translation of these inputs into linked study products. The protocol-interpretation role identifies treatment arms, eligibility requirements, candidate index events, endpoints and unresolved questions. The source-grounding role maps requirements to tables, fields, terminology and join paths. The cohort-base role specifies treatment assignment and time zero. The eligibility role applies condition logic relative to that anchor. The endpoint-analysis role connects the approved cohort to outcome, follow-up, censoring and analysis specifications. In the evaluated implementation, a single language-model agent enacted these roles sequentially using a shared backbone.

Contracts are structured specifications of the inputs, outputs and accepted decisions passed between roles. Inherited decisions include source bindings, patient and encounter identities, treatment groups and time anchors. Treatment assignment and time zero precede baseline eligibility because look-back, washout and follow-up windows depend on the selected anchor [43]. A clarification can revise an upstream decision and trigger regeneration of the affected downstream products.

Cohort construction produces five linked outputs. The patient–group–index base identifies each candidate's group and index event. Condition-level flags record each rule as a binary value for every base patient; final eligibility applies the declared combination of those rules. Single-condition counts summarize each rule against its specified denominator, and cumulative attrition records the ordered cohort funnel. Patient-level and aggregate products are assessed together so that a correct final count cannot conceal an incorrect cohort or condition implementation.

### 4.2.2 Progressive skill selection and exposure

A skill is a versioned procedure for a recurring operation under local data conventions. It combines applicability conditions, source and linkage rules, operational instructions and checks. Examples include distinguishing medication administration from an order, resolving warehouse identifiers to patient identities, interpreting laboratory units and selecting an event-time convention. Protocol-specific thresholds, treatment strategies and estimands are supplied by the study, separately from reusable procedures.

Progressive selection and exposure use the current requirement, workflow stage, accepted source bindings and acquired catalog metadata. The agent first interprets the requirement and discovers candidate sources through catalog search and table-description tools. Source-key and keyword heuristics propose applicable skills, supplemented by model-based description selection when uncertain. The model selection is used when it is valid and nonempty; otherwise, the heuristic selection is retained. Duplicate skills are removed using their identifiers. When no skill is selected, generation can proceed using schema information or an existing SQL draft.

For skills configured for progressive disclosure, the harness supplies sections relevant to base construction or condition-flag generation. Associated reference materials are selected using predefined file-path rules, and larger example collections remain accessible within the document-reading budget. This supplies source-event and identity rules during base construction and condition and time-window rules during flag generation. Contracts provide later roles with accepted source and implementation choices. Explicit feedback updates these choices before the implementation is revised.

This policy selects both the procedure and the portions exposed at a particular decision. The procedural-retrieval comparator instead uses model-generated queries to retrieve passages from the same corpus, with further search and full-document access available. The experimental contrast therefore evaluates the complete selection-and-exposure policy, including content selection, ordering and volume.

### 4.2.3 Validation, revision and human review

Validation checks required fields and types, consistency of identities across products, existence of referenced schema objects, executable output structure and alignment of assignment, baseline and follow-up windows. Execution or interface diagnostics support bounded repair; reference answers are reserved for subsequent scoring. Clinical conflicts and unresolved requirements are referred for clarification. The system records successive versions of study outputs and the earlier decisions and outputs on which each depends.

In institutional use, physicians confirm the research question, treatment strategies, clinical definitions and estimand; analysts review source bindings, joins, observability, cohort attrition and analysis diagnostics. Reviewed corrections or schema changes can inform a new skill version for subsequent studies. Changes to individual studies are documented separately from updates to reusable skills.

After cohort confirmation, the endpoint-analysis role specifies outcomes, follow-up, censoring, covariates and sensitivity analyses. Deterministic libraries carry out propensity-score estimation and overlap assessment [32], covariate-balance assessment [33], Kaplan–Meier and Cox analyses [34,35], and competing-risk analyses where specified [36]. Diagnostic outputs are linked to the reviewed cohort and analysis versions. Institutional export applies the relevant privacy policy.

### 4.3 VTBenchmark construction and reference standards

The benchmark materials, including task requirements and reference annotations, underwent review by two human reviewers. The construction and reference-checking procedures for the two components are described below.

#### 4.3.1 Complex-schema execution component

We constructed 24 protocol roots as the complete crossing of four clinical contexts with six operational challenges. The contexts were oncology, cardiovascular and metabolic disease, renal and chronic disease, and infection and immunity. The challenges were similar source names; multiple sources and operational units; treatment-event status and version; patient-identity granularity and join expansion; event versus documentation time; and eligibility composition with feedback revision. The clinical contexts supplied scenario labels over a shared synthetic semantic structure. The number of roots followed from complete coverage of the four-by-six design.

Each root was represented across four synthetic database environments, yielding 96 task–environment instances. The environments were projected from a common canonical semantic database to vary source selection, identity and linkage structure, temporal interpretation and catalog navigation. These dimensions were motivated by institutional VT implementation and a metadata-only audit of the Mayo Clinic data dictionary (Supplementary Section S1). The constructed schemas represented these operational challenges in synthetic database environments.

The *Simple reference* environment contained 32 relations with clear sources and short paths. *Source ambiguity* contained 128 relations and introduced alternatives such as order, authorization, fulfillment and occurrence evidence, together with patient, encounter, episode, snapshot and history representations. *Temporal and representation ambiguity* retained a comparable catalog scale while increasing ambiguity in event versus documentation time, data grain, status and version. *Large-catalog navigation* contained 1,024 relations, with aliases, bridges and history objects forming paths of two to five hops. The latter was a catalog-navigation stress condition. Reference implementations required only three, four, five and six relations, respectively; the corresponding index-identity paths required one, two, three and four joins (Supplementary Table S1).

Each task specified structured initial requirements and one subsequent clarification replacing an existing upstream requirement, such as the source, status, time anchor or any-versus-all condition logic. Reference computation selected the operational source, applied status and time constraints, resolved enterprise-person identities through the declared linkage path, read current condition assertions and calculated the five products. For the revised phase, the single requirement replacement was applied before recomputation.

Reference checks included recomputation with the same generator and comparison with the prespecified reference products, covering 1,920 product units (24 roots, four environments, two phases, two database versions and five products). Cross-environment checks required agreement across all four physical representations for each root, phase, database role and product, covering 480 consistency units. Prespecified counterexamples distinguished the correct semantics from incorrect sources, clocks, versions and plausible linkage paths that remained executable and returned nonempty results. The same reference rules were applied to both database versions (Section 4.6.1).

### 4.3.2 Source-grounding component

The source-grounding component comprised 90 constructed study tasks based on 55 target-trial emulation publications and 35 institutional virtual-trial intake requests at Mayo Clinic. Publication sources were identified by tracing primary studies cited in the methodological survey by Ren et al. [53] and the systematic review by Hansford et al. [54]. Ren et al. searched PubMed for studies published in top-quartile clinical journals from 1 January 2017 to 15 December 2023 and supplemented these with prior reviews; Hansford et al. searched Medline, Embase, PsycINFO and Web of Science for studies published from March 2012 to October 2022.

Selected references were deduplicated by DOI or normalized title, and bibliographic identifiers were resolved to support full-text retrieval. Retrieved PDFs were parsed using MinerU [55]; parsed text was cleaned before model-assisted extraction of study-design information and structured intake fields. Publications clearly outside the scope of target-trial emulation were excluded during source curation. The benchmark source frame contained 229 entries: 154 external sources and 75 internal intakes drawn from requests submitted between 1 April and 30 June 2026. Subsequent screening retained 208 entries and excluded 21 for duplication within a protocol family or incompatibility with the predefined model input interface. Selection used a deterministic, source-stratified hash ordering of study identifiers, targeting 55 external and 35 internal studies and removing duplicates by DOI, trial registration identifier and protocol family. Coverage characteristics included source type, multi-arm or matched designs, larger sets of conditions, partially unsupported requirements and complex endpoints.

Each task presented the model with structured requirements derived from specified intake materials, treatment-arm and eligibility definitions, supporting text excerpts and structured attributes. Stable requirement identifiers aligned model outputs with references during scoring. The supplied requirements excluded reference table and field mappings, implementation routes and reference disposition labels. Ambiguous scope retained its common or unresolved designation. Missing value sets, terminology, clocks, units or local rules retained a manual-review designation. Tasks were represented in a 1,024-table catalog comprising a curated clinical core of 35 tables and 989 synthetic operational distractors. The clinical core represented local clinical structures described in procedural resources. This component assessed selection of sources for eligibility criteria and treatment arms.

Reference routes were developed through model-assisted proposals and paired model assessments that did not receive one another's responses. A third model assessment used a higher reasoning setting, and necessary structural corrections were made. Two human research leads then confirmed agreement before the reference routes were finalized. Route judgments used the schema, clinical requirement contracts and site-specific procedural descriptions. Required tables captured shared dependencies; each alternative route specified a complete implementation table set, including shared tables. Alternatives conditional on a source system or request interpretation recorded those conditions. Referenced fields had to exist in the supplied schema, and missing local bindings retained a manual-review designation.

Reference answers were constructed without using candidate outputs. Scoring selected among the prespecified acceptable routes. Candidate inputs contained the study requirements, permitted catalog and condition-specific procedural resources; expected routes, reference products and scores were withheld from generation. Clinical task content from internal intake studies remained restricted; the planned public task collection comprises the 55 publication-based studies. The reported evaluation includes all 90 studies.

### 4.4 Comparative evaluation protocols

#### 4.4.1 Four-configuration knowledge experiment

The principal experiment held the harness, language model, stage boundaries, tools, feedback, phase budgets and five-product output requirements constant. Schema-only support supplied the task, schema tools and generic tool instructions. Raw-evidence retrieval-augmented generation supplied data-dictionary entries, field descriptions, relationship metadata, value domains and source documents, excluding procedural modules and worked examples. Shared procedural-corpus retrieval used model-generated queries and BM25 passage retrieval over the admissible corpus. Full VTHarness used progressive selection and exposure over that same corpus (Section 4.2.2). Both procedural configurations had access to the same 12 resources, including six skills, with identical versions.

The matrix crossed 24 roots, four environments, four configurations and two technical repetitions, giving 192 workflows per configuration and 768 overall. Each repetition was a separate end-to-end execution. Every workflow included initial implementation and revision after clarification of a study requirement; full downstream regeneration was allowed within the revision budget. The two prespecified comparisons were full VTHarness versus schema-only support and versus shared procedural-corpus retrieval, pooled across the three complex environments. The Simple reference environment provided a lower-complexity comparison.

#### 4.4.2 Comparisons with adapted systems

We evaluated adapted EmulatRx and HealthFlow [22,23] alongside VTHarness in two separate experiments. EmulatRx uses cohort construction and statistical analysis to support iterative trial design, whereas HealthFlow supports EHR analysis through coordinated agents and reuse of analytical experience. We adapted these workflows to the benchmark tasks and their common five-product delivery requirement. Adaptations connected each system to the assigned inputs, databases and required outputs while retaining its native planning and repair mechanisms. EmulatRx retained its agent graph and shared SQL/Python execution context; HealthFlow retained independent-script execution, evaluation, reflection and within-workflow memory. Both allowed complete regeneration within the phase budgets. During repair, systems received interface and execution diagnostics; reference correctness and final evaluation scores were withheld. The common scoring procedure evaluated the outputs explicitly produced by each system.

The shared-corpus experiment assigned the same 24 roots and two repetitions in the Large-catalog navigation environment to each system, giving 48 workflows per system and 144 overall. All systems could access the same 12 resources, including six procedural skills, through their respective knowledge-use mechanisms; the resource versions were identical. Task bundles, database versions, model parameters and phase budgets were matched. Each HealthFlow workflow began with an empty experience library and retained experience across its initial and revised phases within that workflow. Layered execution, assessable delivery and correctness outcomes were summarized descriptively (Supplementary Section S4).

The separate four-environment experiment crossed 24 roots, four environments and two repetitions for each system, yielding 192 workflows per system and 576 overall. The adapted systems received six metadata, terminology-reference and site-fact resources; VTHarness used its procedural resources and exposure policy. This experiment compared systems with different knowledge inventories (Supplementary Section S5). Both system-comparison experiments and the four-configuration experiment used GPT-5.1 [39], with matched model settings and phase limits within each experiment.

### 4.4.3 Source grounding across language models

We compared the staged harness with and without skills using GPT-5.1 [39], Qwen3-30B-A3B [40] and MedGo [41]. For each backbone, 90 studies, two configurations and three repetitions yielded 540 episodes. Each skill/no-skill pair used identical model settings, episode budgets, task inputs and scoring procedures. Backbones were analyzed separately.

GPT-5.1 runs used temperature 1 and reasoning effort `none`. Source-grounding runs used Qwen3-30B-A3B-Instruct-2507 at temperature 0.7 and MedGo revision `4befd1de86a4` at temperature 0.6, both with thinking disabled. These comparative experiments allowed at most 16,384 output tokens per model response and 100 model calls and 300 tool calls per execution phase or source-grounding episode. MIMIC-IV adaptation used Qwen3-30B-A3B-Instruct-2507 at temperature 1 with a 32,768-token context window and the same output-token ceiling and call limits per task.

## 4.5 Historical-case, database-adaptation and user evaluations

### 4.5.1 Historical Mayo Clinic VT cases

We collected 13 intake–notebook pairs through voluntary sharing of source code by members of the Mayo Clinic VT team. The cases represented new VT requests. The notebooks included manual code revisions made as physicians refined their requests in response to interim findings. They therefore provided records of existing implementations for a descriptive comparison of data-source overlap. The harness received the intake requests, with notebooks withheld, and was evaluated with and without skills using three repetitions per study and configuration.

Requests were matched to notebooks using intake identifiers, and file hashes were used to verify the collected versions. Prespecified rules defined eligible cohort-code cells and fields; endpoint, follow-up and outcome code were outside this scope. References were extracted from code rather than saved notebook outputs. Table references were extracted by whole-word matching of data-dictionary table names in the retained code, including comments. Primary field references were extracted from SQL literals by parsing BigQuery syntax, resolving aliases within query scopes and validating table–field pairs against the schema. Parsing failures and ambiguous column assignments were recorded as unresolved.

We evaluated the tables explicitly selected at the end of source selection. Fields were evaluated from the recorded selections before the corresponding full table schemas were added. Validation checked coverage of candidate requirements, requirement-layer consistency, agreement of declared table sets and field membership in the displayed schema. Table and field recall measured overlap between the explicit selections and the corresponding references extracted from analyst notebooks (Methods, Section 4.6). Requirement representation was evaluated separately through author-adjudicated semantic alignment.

### 4.5.2 Adaptation to MIMIC-IV

We evaluated adaptation to a different database and task set using eight cohort-construction tasks on MIMIC-IV 3.1 [42]. A deterministic hash ordering selected 1,000 subjects before outcome inspection. Their records were retained in nine tables, preserving original identities, units and full shifted timestamps. Tasks combined adult age at the selected admission with demographic, laboratory, ICU, discharge or medication-administration predicates (Supplementary Table S10).

Local identity, event-time, unit and administration conventions were encoded in five MIMIC-IV skills with nine parameterized operations. Task-specific thresholds and literals were supplied separately. Site procedures and runtime interfaces were refined during adaptation using the eight reported tasks and the same patient-data subset. After adaptation, each task was run once with Qwen3-30B-A3B-Instruct-2507.Separate SQL and pandas reference implementations agreed on all tasks and synthetic counterexamples; reference implementations and outputs were withheld from generation. Success required execution and exact delivery of all five products. Skill formation, task predicates and checking procedures are detailed in Supplementary Section S8.

### 4.5.3 Institutional application and analyst questionnaire

We assessed the Mayo Clinic application by reviewing study outputs and analysis documentation to identify the supported workflow capabilities. The assessment covered continuity from protocol interpretation and cohort construction to propensity-score and balance diagnostics, survival and competing-risk analysis, investigator review and controlled export. Reporting focused on the supported operations and their connections across workflow stages (Supplementary Section S10).

Five analysts with prior workflow experience completed a structured questionnaire covering nine five-point ratings, useful components and paired estimates of active work with and without VTHarness. Each pair referred to the same scenario and assumed immediate physician clarification. The unnecessary-complexity item had the opposite favorable direction to the other ratings. Responses were de-identified and summarized at respondent level.

## 4.6 Outcome definitions and statistical analysis

### 4.6.1 Execution and complete delivery

Each synthetic database had an original version for generating and debugging candidate programs and a test version with consistently relabeled patient and relational identifiers. Both versions shared the same schema, clinical event values, study rules and reference counts. Systems could inspect and query the original version during generation and debugging. The generated program and its dependencies were then run unchanged on the relabeled version, with no further model calls or edits. All systems used the same versions within each environment.

Program execution measured successful execution of the selected program on the relabeled version. Five-product assessability required all five explicit outputs to be readable and mappable to predefined cohort structures on both versions. For the adapted systems, empty base cohorts were classified as unassessable. Native and adapted systems used predefined, system-specific projections to select semantic columns, resolve declared bindings, order records and parse flags and counts. In the synthetic execution experiments, index values were compared using the first ten characters of their date strings. Projections for the adapted systems retained identifier value types and explicit group labels; native exports used their candidate-side group bindings. After these transformations, each output was compared with its reference using exact object equality, with no numerical tolerance. Extra output columns generally did not enter this comparison.

Complete correct delivery (hereafter, complete delivery) required all five transformed outputs to match their references on both versions. The system also had to complete its execution phase and run the unchanged program successfully on the relabeled version. Execution status was assessed separately from product content. Scoring used the submitted outputs as generated, with missing required products or fields counted as incomplete delivery. Revision outcomes distinguished retained correctness, recovery, regression and persistent failure.

All assigned workflows remained in the denominators. Non-delivery, budget exhaustion, invalid outputs, execution failure and incorrect products therefore contributed to incomplete delivery. Resource summaries counted recorded model and tool calls, including failed attempts. Token-use summaries were based on available records.

A post hoc descriptive audit of the four-configuration experiment used archived outputs and prespecified references to distinguish complete export from correct cohort construction. For revised workflows in the three complex environments, it recorded whether all five products were produced on both versions and whether either base cohort was empty despite a nonempty reference. For the shared-corpus comparison, a separate audit examined execution records, generated outputs, assessability decisions and reference comparisons for each workflow and phase. It checked the nesting of the three reported indicators and examined selected cases to characterize implementation errors.

### 4.6.2 Source recovery and user outcomes

Source-grounding outputs were aligned to references by exact matching of the predefined requirement identifiers. Unknown or duplicate retrieval identifiers and changes to requirement layer or polarity were rejected.

For each aligned requirement, $E$ denoted an acceptable expected-table route and $A$ the candidate table set. We defined $TP = |A \cap E|$, $FN = |E \setminus A|$ and $FP = |A \setminus E|$, and calculated

$$\text{Recall} = \frac{TP}{TP + FN}, \qquad \text{F1} = \frac{2TP}{2TP + FP + FN}, \qquad \text{F2} = \frac{5TP}{5TP + 4FN + FP}.$$

When multiple prespecified routes were acceptable, selection maximized recall, then F2, then precision, and finally minimized the number of extra tables. Remaining ties retained the first route in the prespecified order: nonempty required tables preceded listed alternatives, with duplicate table sets removed in order. Complete-route coverage indicated recovery of all expected tables in the selected route. For an aligned candidate with an empty reference set, recall and complete-route coverage were one; F1 and F2 were one when the candidate set was also empty and zero when it contained extra tables. Failed outputs or the absence of a scoreable aligned candidate received zero recall, F1, F2 and complete-route credit.

Criterion and treatment-arm requirements formed one scoring set. Each reference requirement ordinarily received weight one. An indistinguishable group of $k$ reference identifiers with the same route received weight $1/k$ per identifier, giving the group total weight one. Identity-indistinguishable groups with different routes received zero weight and were excluded from the point-estimate denominator. Each metric was calculated per requirement and then averaged using these reference weights within an episode. For study $s$, configuration $c$, repetition $e$ and requirement $r$, with reference weight $w_{sr}$ and requirement score $m_{secr}$, aggregation was

$$M_{sec} = \frac{\sum_r w_{sr}\, m_{secr}}{\sum_r w_{sr}}, \qquad M_{sc} = \frac{1}{3}\sum_{e=1}^{3} M_{sec}, \qquad \overline{M}_c = \frac{1}{90}\sum_{s=1}^{90} M_{sc}.$$

Thus, weighting operated at the requirement level within episodes and equally across studies after averaging repetitions. Unconditional expected-table recall was the primary source-grounding outcome; F1, F2 and complete-route recovery were additional outcomes. Precision was used in route selection. Delivery-conditioned summaries described the quality of returned outputs.

For historical cases, table recall was the proportion of notebook-reference tables that appeared in the system's valid explicit selections. Field recall was the proportion of reference fields that appeared in the system's explicit selections. Reference fields were schema-validated, table-qualified fields extracted from parseable notebook SQL. Missing final source outputs received zero credit; field recall remained undefined when the reference field set was empty. Three repetitions were averaged within each study and configuration, followed by equal weighting across studies. These outcomes measured overlap between sources explicitly selected by the system and references extracted from existing analyst code.

The separate auxiliary requirement-representation endpoint used author-confirmed semantic alignments after generation, with model recommendations informing adjudication. Unresolved or unassessable requirements received no credit, and missing outputs remained in the denominator. Questionnaire outcomes comprised rating distributions, component-selection frequencies and paired estimates of active work. Hours saved and percentage reduction were calculated within respondent before reporting their medians.

### 4.6.3 Statistical analysis

We estimated paired differences and nonparametric bootstrap intervals [56], preserving the unit of replication for each evaluation. In the four-configuration experiment, two technical repetitions were first averaged within each root–environment–configuration combination. For each of 24 roots, configuration differences were then averaged across the three complex environments. A paired percentile bootstrap resampled roots with all their environments and configurations together, using 10,000 draws. Bonferroni-adjusted 97.5% marginal intervals accounted for the two prespecified comparisons. Environment-specific and revision-transition results were summarized descriptively.

The separate four-environment system-comparison experiment used the same root-level averaging and paired resampling structure across all four environments, with 100,000 draws and 97.5% marginal intervals for its two prespecified comparisons. The shared-corpus comparison reported descriptive outcomes across 48 assigned workflows per system.

For each backbone in the 90-study source-grounding evaluation, 10,000 paired bootstrap samples resampled studies after averaging repetitions within study. The historical-case analysis similarly resampled the 13 studies. For the questionnaire, 10,000 bootstrap samples resampled the five respondents to estimate uncertainty in the median paired hours saved. The eight-task MIMIC-IV evaluation was summarized descriptively. Intervals were two-sided 95% intervals unless specified otherwise. The analysis emphasized effect estimates and uncertainty.

## 5 Data availability

Following completion of the applicable Mayo Clinic institutional review and receipt of approval, the public components of VTBenchmark will be made available in a public repository. The release will comprise the 24 complex-schema roots across four synthetic environments, the corresponding synthetic databases and reference products, and the 55 publication-based source-grounding tasks with their catalog and references. Reported source-grounding estimates include all 90 evaluated studies; the 35 internal intake-based tasks will not be publicly distributed. Institutional dictionaries, intake forms, analyst notebooks, clinical records and estimates, patient-derived request archives and questionnaire microdata will remain restricted because they contain institutional, clinical or participant-related information. MIMIC-IV 3.1 is available through its established access process [42]. Repository and archival identifiers for the VTBenchmark release will be provided after Mayo Clinic approval.

## 6 Code availability

Following review by the Mayo Clinic and receipt of the required institutional approval, the custom code and related AI artifacts supporting this study will be publicly released through GitHub, as appropriate. The release will be limited to the materials authorized by Mayo Clinic and will comply with any licensing, access, use and timing conditions specified in the approval. Repository links, archival identifiers and license terms will be provided after approval.

## Acknowledgements

The authors acknowledge the Mayo Clinic Virtual Trial team and the institutional research computing and data governance infrastructure that supported this work. This work was supported by the National Institutes of Health under grants R01AG084236 and R01AG083039, and by the Nancy Peretsman and Robert Scully Endowed Chair in Artificial Intelligence and Informatics at Mayo Clinic.

## Author contributions

**Pengze Li:** Conceptualization, Methodology, Software, Validation, Formal analysis, Investigation, Data curation, Visualization, Project administration, Writing–original draft, and Writing–review & editing. **Jingna Feng:** Methodology, Validation, Investigation, and Writing–review & editing. **Yue Yu:** Methodology, Validation, Investigation, and Writing–review & editing. **Yining Wu:** Data curation, Visualization, and Writing–review & editing, including benchmark construction and figure preparation. Jingna Feng, Yue Yu and Xinyue Hu contributed virtual trial-execution expertise, guidance on the Mayo Clinic real-world clinical data environment, and technical review from data and trial-design perspectives. **Avanti Bhandarkar:** Conceptualization and Methodology related to the agent architecture. **Xinyue Hu:** Investigation, Validation, and Writing–review & editing. **Jie Lin:** Investigation, Validation, and Writing–review & editing. Jingna Feng, Yue Yu, Xinyue Hu, Avanti Bhandarkar and Jie Lin tested the system and provided feedback that informed iterative improvements. **Cui Tao:** Conceptualization, Methodology, Supervision, Project administration, Resources, Funding acquisition, and Writing–review & editing.

## Competing interests

The authors declare no competing interests.